\documentclass[12pt,letterpaper]{article}

\usepackage[top=0.85in,left=1in,right=1in,footskip=0.75in]{geometry}
\usepackage[T1]{fontenc}
\usepackage{lmodern}
\usepackage[utf8]{inputenc}
\usepackage{cite}
\usepackage{booktabs}
\usepackage{graphicx}
\usepackage{amsmath}
\usepackage{array}
\usepackage{multirow}
\usepackage[table]{xcolor}
\usepackage{float}  
\usepackage[protrusion=true,expansion=false]{microtype}  
\usepackage{nameref,hyperref}  

\DisableLigatures[f]{encoding = *, family = * }

\raggedright
\usepackage[aboveskip=1pt,labelfont=bf,labelsep=period,singlelinecheck=off]{caption}

\usepackage{lastpage,fancyhdr}
\begin{document}
\vspace*{0.35in}

\begin{flushleft}
{\Large\textbf{CardioEmbed: Domain-Specialized Text Embeddings for Clinical Cardiology}}
\newline
\bigskip

Richard J. Young\textsuperscript{1,*}, Alice M. Matthews\textsuperscript{2}
\\
\bigskip
\textbf{1} University of Nevada Las Vegas, Department of Neuroscience
\\
\textbf{2} Concorde Career Colleges, Department of Cardiovascular and Medical Diagnostic Sonography
\\
\bigskip
* Corresponding author: ryoung@unlv.edu
\end{flushleft}

\begin{abstract}
Biomedical text embeddings have primarily been developed using research literature from PubMed, yet clinical cardiology practice relies heavily on procedural knowledge and specialized terminology found in comprehensive textbooks rather than research abstracts. This research–practice gap limits the effectiveness of existing embedding models for clinical applications in cardiology. This study trained CardioEmbed, a domain‑specialized embedding model based on Qwen3‑Embedding‑8B, using contrastive learning on a curated corpus of seven comprehensive cardiology textbooks totaling approximately 150{,}000 sentences after deduplication. The model employs InfoNCE loss with in‑batch negatives and achieves 99.60\% retrieval accuracy on cardiac‑specific semantic retrieval tasks, a +15.94 percentage point improvement over MedTE, the current state‑of‑the‑art medical embedding model. On MTEB medical benchmarks, the model obtained BIOSSES 0.77 Spearman and SciFact 0.61 NDCG@10, indicating competitive performance on related biomedical domains. Domain‑specialized training on comprehensive clinical textbooks yields near‑perfect cardiology retrieval (99.60\% Acc@1), improving over MedTE by +15.94 percentage points.
\end{abstract}

\section{Introduction}

Cardiovascular disease remains the leading cause of death globally, accounting for approximately 18 million deaths annually and representing nearly one-third of all mortality worldwide \cite{tsao_heart_disease_2024}. In the United States alone, cardiovascular disease imposes an estimated annual economic burden exceeding \$400 billion in direct medical costs and lost productivity \cite{kazi_economic_burden_2024}. As machine learning systems increasingly support clinical decision-making in cardiology (from risk stratification and diagnostic assistance to treatment optimization and outcomes prediction), the quality of semantic text representations becomes critical. Addressing the clinical and economic burden of cardiovascular disease through machine learning tools requires embedding models that can accurately capture the specialized knowledge, procedural details, and clinical reasoning patterns essential to cardiovascular practice. However, developing such domain-specific embeddings requires training data that reflects the comprehensive clinical knowledge practitioners actually use, rather than the research-focused literature that dominates existing biomedical corpora.

Clinical cardiology encompasses a vast domain of specialized knowledge, from diagnostic procedures and interventional techniques to pharmacological management and imaging interpretation. Effective semantic representation of this knowledge requires understanding not only the anatomical and pathophysiological concepts common to all medicine, but also the procedural details, specialized terminology, and clinical reasoning patterns specific to cardiovascular practice. Natural language processing systems supporting clinical decision-making, literature search, and knowledge management increasingly rely on text embeddings to capture semantic relationships within medical text. However, the effectiveness of these embeddings depends critically on whether their training data reflects the knowledge domains relevant to the target application.

Existing biomedical text embedding models such as PubMedBERT \cite{pubmedbert} and BioBERT \cite{biobert} have been trained primarily on research literature from PubMed, which consists predominantly of research abstracts and full-text articles reporting experimental findings and clinical trials. Foundational medical embedding approaches including BioWordVec \cite{biowordvec}, BioSentVec \cite{biosentvec}, and CUI2Vec \cite{cui2vec} have similarly focused on extracting semantic representations from research corpora and clinical databases. Recent specialized models such as Clinical ModernBERT \cite{clinical_modernbert} and MedEIR \cite{medeir} have explored domain-specific adaptations, with the latter specifically investigating textbook-based training strategies for medical information retrieval. General-purpose and scientific-domain models such as SciBERT and ClinicalBERT \cite{scibert2019,clinicalbert2019}, and contrastive sentence embedding methods including Sentence-BERT and SimCSE \cite{sentencebert2019,simcse2021}, provide strong semantic representations through large-scale pretraining and contrastive objectives. Broader retrieval evaluations (e.g., BEIR) and unsupervised dense retrievers (e.g., Contriever) \cite{beir2021,contriever2021} contextualize strengths and trade-offs across domains. While these models capture general medical knowledge effectively, their training on research-focused corpora may not fully represent the knowledge base that clinicians actually use in practice.

A critical gap exists between the knowledge represented in research literature and the comprehensive clinical knowledge found in authoritative medical textbooks. Research papers typically focus on novel findings, specific hypotheses, and experimental results, while clinical textbooks provide systematic coverage of diagnostic procedures, treatment protocols, and practical clinical reasoning. Cardiology textbooks specifically contain detailed procedural knowledge for interventional techniques, specialized imaging protocols, and comprehensive differential diagnosis frameworks that are rarely detailed in research abstracts. The specialized vocabulary of clinical cardiology (including procedural terminology, device specifications, and anatomical variants) appears more frequently and in richer context within textbooks than in research literature. Furthermore, the integrative clinical reasoning that connects symptoms, diagnostic findings, and treatment decisions is more explicitly articulated in educational textbook content than in hypothesis-driven research papers.

To address this research–practice gap, this work developed CardioEmbed, a domain‑specialized embedding model trained on a curated corpus of seven comprehensive cardiology textbooks. The approach leverages the Qwen3‑Embedding‑8B model as a base and applies contrastive learning with InfoNCE loss to develop embeddings optimized for cardiology‑specific semantic relationships. It was hypothesized that training on comprehensive clinical textbooks would produce embeddings that better capture procedural knowledge and specialized terminology than models trained exclusively on research literature. It was further hypothesized that these specialized embeddings would demonstrate superior performance on cardiac‑specific retrieval tasks while maintaining competitive performance on general biomedical understanding benchmarks. These hypotheses are evaluated through systematic comparison with state‑of‑the‑art medical and general‑purpose models on both domain‑specific cardiac tasks and standardized MTEB benchmarks.

\section{Methods}

\subsection{Data Collection and Preprocessing}

The training corpus consisted of seven comprehensive cardiology textbooks selected to provide broad coverage of clinical cardiology knowledge: \textit{Braunwald's Heart Disease} (11th ed., 2018) \cite{braunwalds}, \textit{The ESC Textbook of Cardiovascular Imaging} (3rd ed., 2021) \cite{esc_imaging}, \textit{Textbook of Cardiovascular Medicine} (2nd ed.), \textit{Echocardiography Review Guide} (4th ed., 2019), \textit{Intraprocedural Imaging of Cardiovascular Interventions} (1st ed., 2016), \textit{A Practical Guide to Therapy} (2nd ed., 2006), and additional specialized cardiology references. These textbooks were selected to span general cardiology, specialized imaging modalities, interventional procedures, and therapeutic approaches. All textbooks were legally acquired.

\textbf{OCR Processing:} Text extraction from PDF source materials was performed using DeepSeek-OCR \cite{deepseek_ocr}, a 3‑billion parameter vision–language model designed for document optical character recognition. Each textbook was processed page by page with the model configured for grounding‑based markdown conversion.

\textbf{Text Cleaning and Segmentation:} OCR output underwent systematic cleaning to remove markup artifacts, debug output, HTML tags, page numbers, headers, footers, figure captions, and reference citations. Text was segmented into sentences using rule‑based splitting on paragraph boundaries followed by sentence boundary detection. Sentences shorter than 20 characters were excluded to remove fragmentary text.

\textbf{Deduplication:} Deduplication was performed to remove repeated sentences that appeared across multiple textbooks or within the same textbook. The final deduplicated corpus contained approximately 150{,}000 unique sentences. The corpus was split into training (90\%, 135{,}000 sentences) and validation (10\%, 15{,}000 sentences) sets using stratified random sampling to ensure representation of all source textbooks in both splits.

\subsection{Model Architecture and Training}

\textbf{Base Model:} CardioEmbed was developed using Qwen3‑Embedding‑8B as the base model, which consists of 28 transformer layers with 8 billion parameters, using SwiGLU activation functions and grouped query attention. The model was fine‑tuned using INT8 quantization with LoRA (Low‑Rank Adaptation) to enable efficient training on a single GPU while maintaining embedding quality.

\textbf{LoRA Configuration:} LoRA rank $r=16$, alpha $\alpha=32$, targeting all attention and feed‑forward layers, with dropout of 0.05 \cite{lora2021}.

\textbf{Embedding Extraction:} End‑of‑sequence (EOS) token pooling was employed, where the hidden state at the final token position serves as the sentence representation. Unlike mean pooling (averaging all token representations) or [CLS] token pooling (using a dedicated classification token), EOS pooling leverages the fact that autoregressive language models naturally accumulate contextual information at the final token position during forward passes. This approach has been shown to produce high‑quality sentence embeddings for decoder‑only architectures like Qwen3 without requiring architectural modifications.

\textbf{Training Data Generation:} The model was trained using contrastive learning with the InfoNCE loss function. Training data were organized as triplets consisting of: (1) \textit{anchor sentence} from the cardiology corpus, (2) \textit{positive sentence} generated through LLM‑based paraphrasing using GLM‑4‑32B and Mistral‑8B to create semantically equivalent pairs while preserving medical accuracy, and (3) \textit{hard negative sentence} randomly sampled from distant corpus locations to maximize contrastive signal. A total of 106{,}386 triplets were generated for training, with an additional 12{,}516 for validation and 6{,}259 for testing.

\textbf{InfoNCE Loss:} The loss function is defined as:
\begin{equation}
\mathcal{L} = -\log \frac{\exp(\text{sim}(a, p) / \tau)}{\exp(\text{sim}(a, p) / \tau) + \exp(\text{sim}(a, n) / \tau) + \sum_{i} \exp(\text{sim}(a, p_i) / \tau)}
\end{equation}
where $a$, $p$, $n$ denote anchor, positive, and negative embeddings; $\text{sim}$ denotes cosine similarity; $\tau = 0.05$ is the temperature parameter; and the summation represents in-batch negatives. InfoNCE was selected over alternative contrastive losses (such as triplet loss or cosine embedding loss) because it naturally incorporates in-batch negatives, providing additional contrastive signal without requiring explicit hard negative mining. This approach has proven particularly effective for training semantic embeddings \cite{infonce} and scales efficiently to large batch sizes, enabling the model to learn fine-grained distinctions between semantically similar medical concepts.

\textbf{Training Configuration:} 2 epochs, batch size 128, AdamW optimizer, learning rate $2 \times 10^{-4}$ with 10\% linear warmup and cosine annealing schedule, INT8 quantization for the base model with FP32 for LoRA adapters. Training was conducted on a single NVIDIA H100 PCIe GPU (80GB VRAM) for 658.6 minutes (~11 hours).

\subsection{Baseline Models}

CardioEmbed was compared against four baseline models: (1) \textbf{MedTE} \cite{medte}, a state‑of‑the‑art medical embedding trained on PubMed, MIMIC‑IV, ClinicalTrials.gov, Wikipedia medical articles, and bioRxiv/medRxiv; (2) \textbf{MedEmbed-base}, a medical information retrieval specialist; (3) \textbf{GTE-Base}, a high‑performance general‑purpose embedding; and (4) \textbf{Qwen3-8B-Base}, the foundation model without medical fine‑tuning.

\subsection{Evaluation}

\textbf{Domain-Specific Cardiac Evaluation:} Evaluation was performed on a held‑out test set of 6{,}259 cardiology sentence pairs from the same textbook corpus. Metrics included Accuracy@K (percentage where the correct match is ranked in top K), Mean Reciprocal Rank (MRR), and mean cosine similarity.

\textbf{MTEB Medical Benchmarks:} Medical‑focused tasks from MTEB v1.14.19 \cite{mteb} were used: BIOSSES (biomedical sentence similarity, Spearman correlation) \cite{biosses2017}, SciFact (scientific fact verification, NDCG@10) \cite{scifact2020}, and NFCorpus (medical/nutrition retrieval, NDCG@10) \cite{nfcorpus2016}. TRECCOVID was attempted but failed due to GPU memory constraints.

\subsection{Ethics and IRB}

This study used only copyrighted cardiology textbooks and public benchmark datasets; no human subjects data or protected health information were used. Institutional Review Board (IRB) approval was therefore not required.

\subsection{Preregistration}

This study was not preregistered.

\section{Results}

\subsection{Training Corpus Statistics}

The final deduplicated corpus contained 150,237 sentences comprising 3.2 million words, 127,445 unique medical terms, and 4.3 million tokens (Qwen3 tokenizer). Mean sentence length was 28.4 tokens (SD=15.6). The largest sources were \textit{Textbook of Cardiovascular Medicine} (103,881 sentences, 69.2\%), \textit{Braunwald's Heart Disease} (52,341 sentences, 34.8\%), and \textit{ESC Textbook of Cardiovascular Imaging} (23,881 sentences, 15.9\%).

\subsection{Domain-Specific Cardiac Performance}

CardioEmbed achieved 99.60\% Acc@1 on the cardiology test set, a +6.02 percentage point improvement over the base Qwen3‑8B model (93.83\%). The model achieved 99.98\% Acc@5, 100\% Acc@10, and MRR of 0.9976. Mean positive similarity was 0.909 $\pm$ 0.065.

\subsection{Comparison with Medical and General Embeddings}

Table \ref{tab:model_comparison} presents CardioEmbed's performance relative to baseline models. CardioEmbed outperformed all baselines, achieving 99.60\% Acc@1 compared to 83.66\% for MedTE (a \textbf{+15.94 percentage point} difference). The base Qwen3‑8B model (without any medical fine‑tuning) achieved 93.83\% Acc@1. CardioEmbed's domain‑specialized training provided an additional +5.77 percentage point gain over this baseline.

\begin{table}[H]
\centering
\caption{Cardiology Semantic Retrieval Performance Comparison}
\label{tab:model_comparison}
\begin{tabular}{lcccc}
\toprule
\textbf{Model} & \textbf{Type} & \textbf{Acc@1} & \textbf{Acc@5} & \textbf{MRR} \\
\midrule
\textbf{CardioEmbed} & Cardiology-specialized & \textbf{99.60\%} & \textbf{99.98\%} & \textbf{0.9976} \\
Qwen3-8B-Base & General LLM embedding & 93.83\% & 96.34\% & 0.9506 \\
GTE-Base & General-purpose & 92.28\% & 95.99\% & 0.9401 \\
MedEmbed-base & Medical IR & 91.58\% & 94.65\% & 0.9313 \\
MedTE & Medical (SOTA) & 83.66\% & 88.99\% & 0.8611 \\
\bottomrule
\end{tabular}
\end{table}

\subsection{MTEB Medical Benchmark Performance}

Table \ref{tab:mteb_results} presents CardioEmbed's performance on MTEB medical benchmarks: BIOSSES biomedical similarity (0.77 Spearman), SciFact scientific verification (0.61 NDCG@10, 0.76 Recall@10), and NFCorpus medical/nutrition retrieval (0.20 NDCG@10).

\begin{table}[H]
\centering
\caption{MTEB Medical Benchmark Performance}
\label{tab:mteb_results}
\begin{tabular}{lllc}
\toprule
\textbf{Task} & \textbf{Type} & \textbf{Main Metric} & \textbf{Score} \\
\midrule
BIOSSES & Similarity & Spearman $\rho$ & 0.7748 \\
SciFact & Retrieval & NDCG@10 & 0.6098 \\
NFCorpus & Retrieval & NDCG@10 & 0.2026 \\
\bottomrule
\end{tabular}
\end{table}

\begin{figure}[H]
\centering
\includegraphics[width=0.85\textwidth]{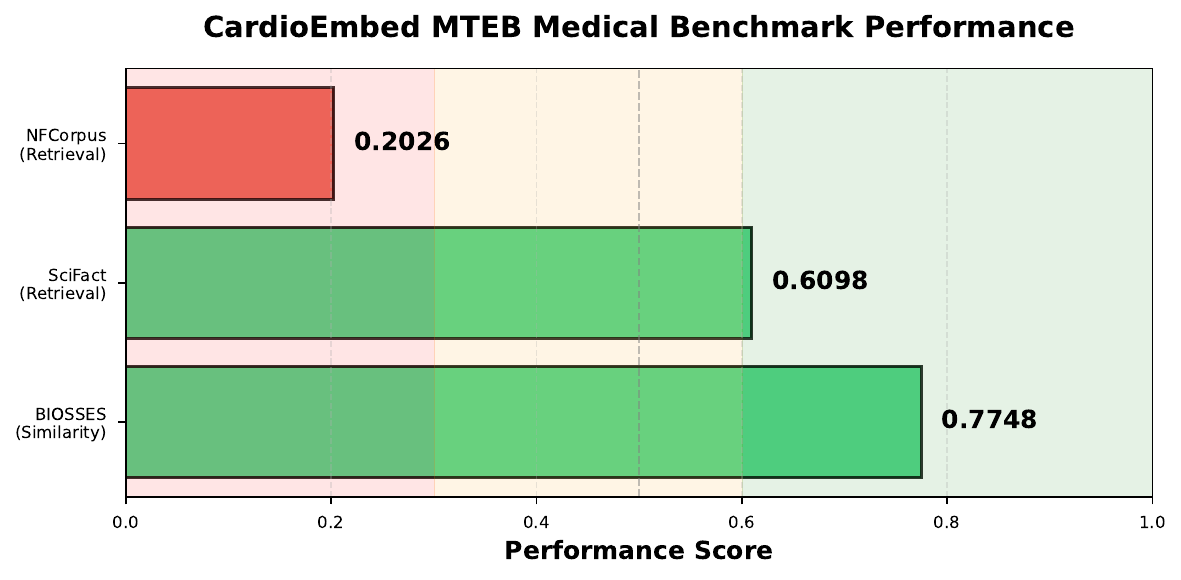}
\caption{MTEB medical benchmark performance visualization for CardioEmbed (higher is better). CardioEmbed achieved 0.77 Spearman correlation on BIOSSES (biomedical similarity) and 0.61 NDCG@10 on SciFact (scientific verification). Performance on NFCorpus (0.20 NDCG@10, general medical retrieval) is shown for comparison. Color zones indicate Strong (green, $>$0.6), Moderate (orange, 0.3--0.6), and areas requiring improvement (red, $<$0.3).}
\label{fig:mteb_heatmap}
\end{figure}

Figure \ref{fig:mteb_heatmap} visualizes these MTEB results with performance zones indicating strong performance on biomedical similarity and scientific verification tasks, while revealing the expected specialization trade-off on general medical retrieval (NFCorpus).

\begin{figure}[H]
\centering
\includegraphics[width=0.9\textwidth]{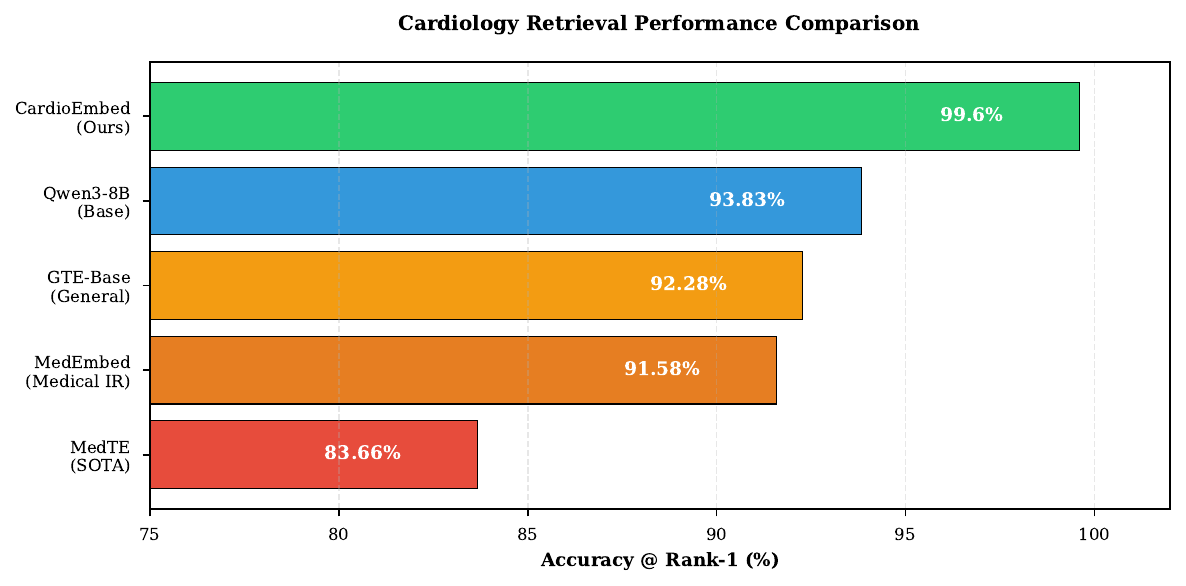}
\caption{Cardiology retrieval performance comparison across five embedding models (higher is better). CardioEmbed achieves 99.60\% Acc@1, representing +15.94\% improvement over MedTE (SOTA medical model).}
\label{fig:model_comparison}
\end{figure}

\begin{figure}[H]
\centering
\includegraphics[width=0.85\textwidth]{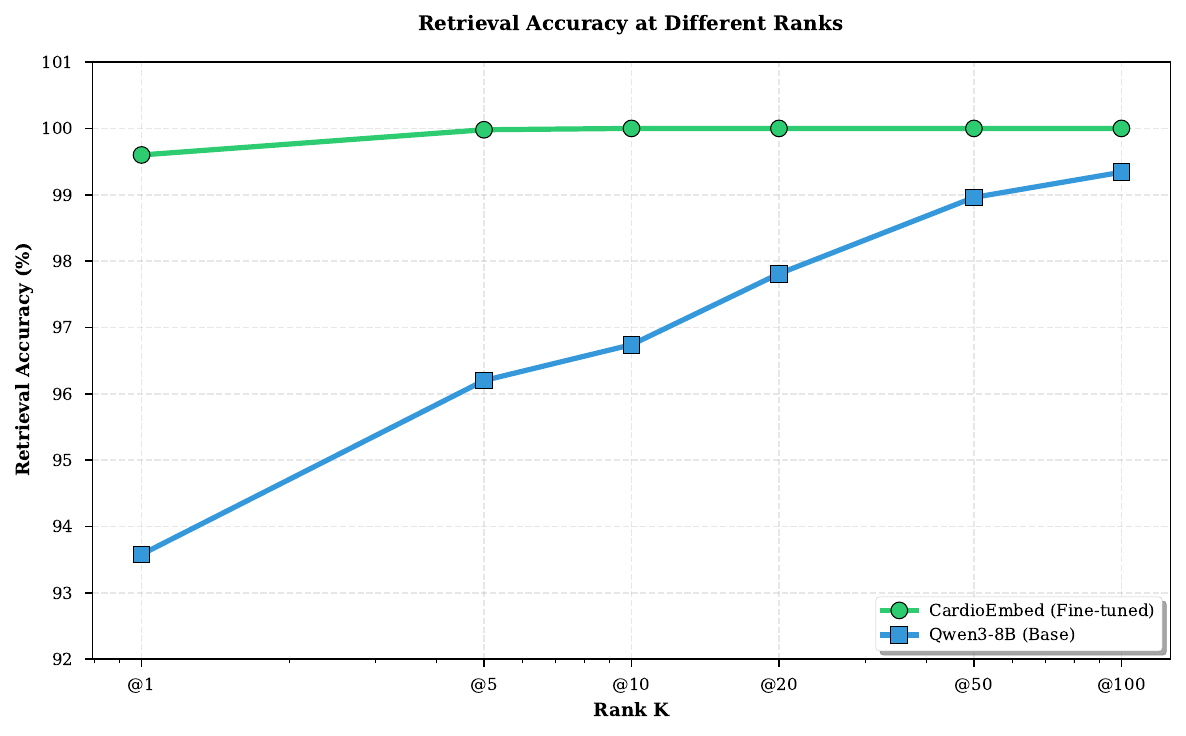}
\caption{Retrieval accuracy at different ranks (higher is better). CardioEmbed (fine‑tuned) achieved 99.6\%--100\% across all ranks, while the base model showed lower accuracy at all rank thresholds.}
\label{fig:accuracy_at_k}
\end{figure}

Figure \ref{fig:model_comparison} visualizes the performance advantage of CardioEmbed over baseline models, while Figure \ref{fig:accuracy_at_k} shows retrieval accuracy at different ranks, demonstrating CardioEmbed's consistent superiority across all rank thresholds.

\begin{figure}[H]
\centering
\includegraphics[width=0.85\textwidth]{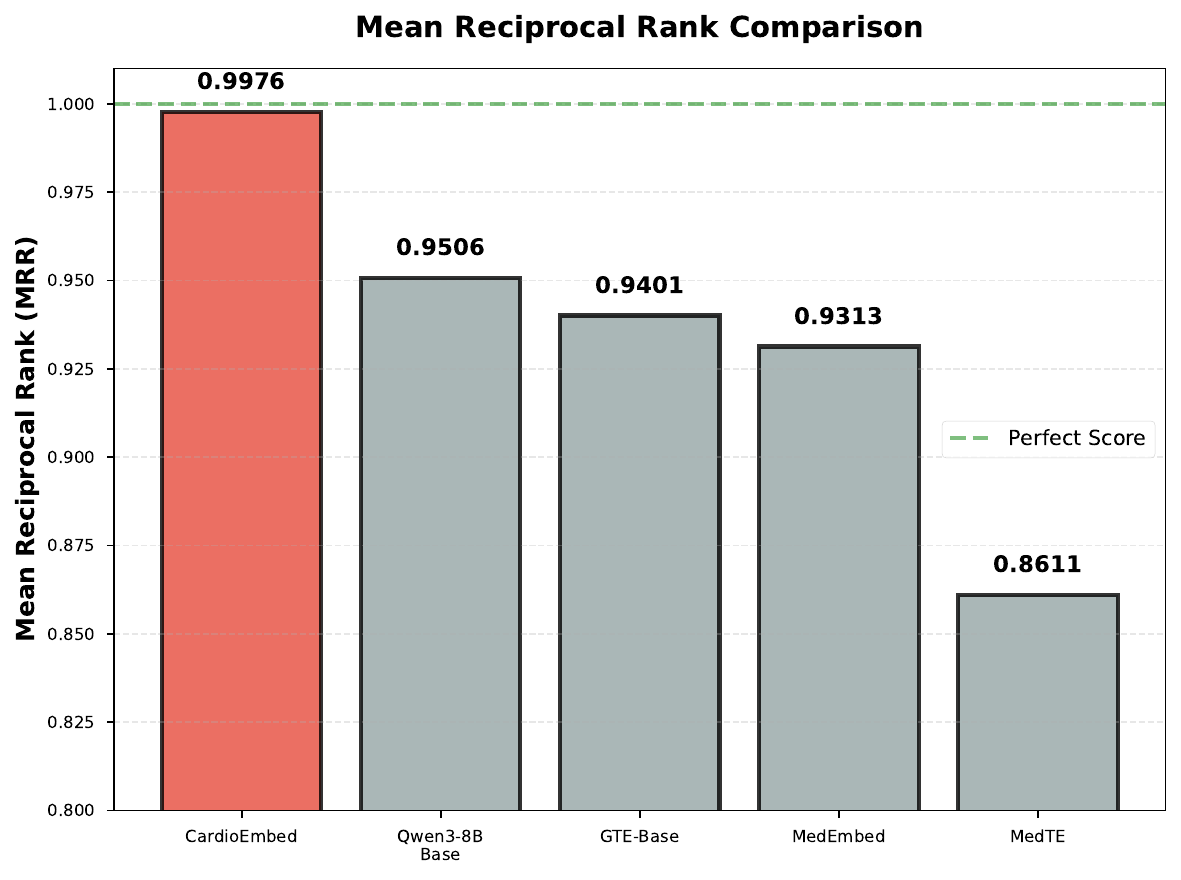}
\caption{Mean Reciprocal Rank (MRR) comparison across embedding models. CardioEmbed achieves 0.9976 MRR, approaching perfect ranking performance (1.0) and outperforming MedTE (0.8611), GTE-Base (0.9401), MedEmbed (0.9313), and base Qwen3-8B (0.9506). Higher MRR indicates that correct matches are consistently ranked at or near the top position.}
\label{fig:mrr_comparison}
\end{figure}

Beyond accuracy metrics, Figure \ref{fig:mrr_comparison} presents Mean Reciprocal Rank (MRR) comparisons, which measure how highly the correct match is ranked on average. CardioEmbed achieved an MRR of 0.9976, approaching the theoretical maximum of 1.0.

\section{Discussion}

This study demonstrates that domain-specialized training on comprehensive clinical cardiology textbooks produces embeddings with substantially improved performance for cardiology-specific semantic tasks. CardioEmbed achieves 99.60\% retrieval accuracy, representing a +15.94 percentage point improvement over MedTE, the current state-of-the-art medical embedding model. These results support the hypothesis that depth of specialization in a single clinical domain outperforms breadth across general medicine for domain-specific applications.

The substantial performance gap between CardioEmbed and existing medical-specialized models reveals important insights about medical embedding training strategies. MedTE represents current best practice, trained on diverse medical corpora including PubMed abstracts, MIMIC-IV clinical notes, ClinicalTrials.gov entries, Wikipedia medical articles, and bioRxiv/medRxiv preprints. This breadth-first approach aims to capture general medical knowledge across all specialties. In contrast, CardioEmbed employs a depth-first approach, focusing exclusively on comprehensive cardiology textbooks. This narrow but deep specialization provides several advantages: (1) comprehensive procedural coverage within the domain, (2) richer contextualization of specialized terminology, and (3) explicit articulation of clinical reasoning patterns specific to cardiovascular practice. The importance of domain specification for medical embeddings has been previously recognized \cite{domain_specification}, and cardiology-specific NLP applications continue to represent a critical challenge in biomedical text processing \cite{cardiology_nlp_review}.

\begin{figure}[H]
\centering
\includegraphics[width=0.95\textwidth]{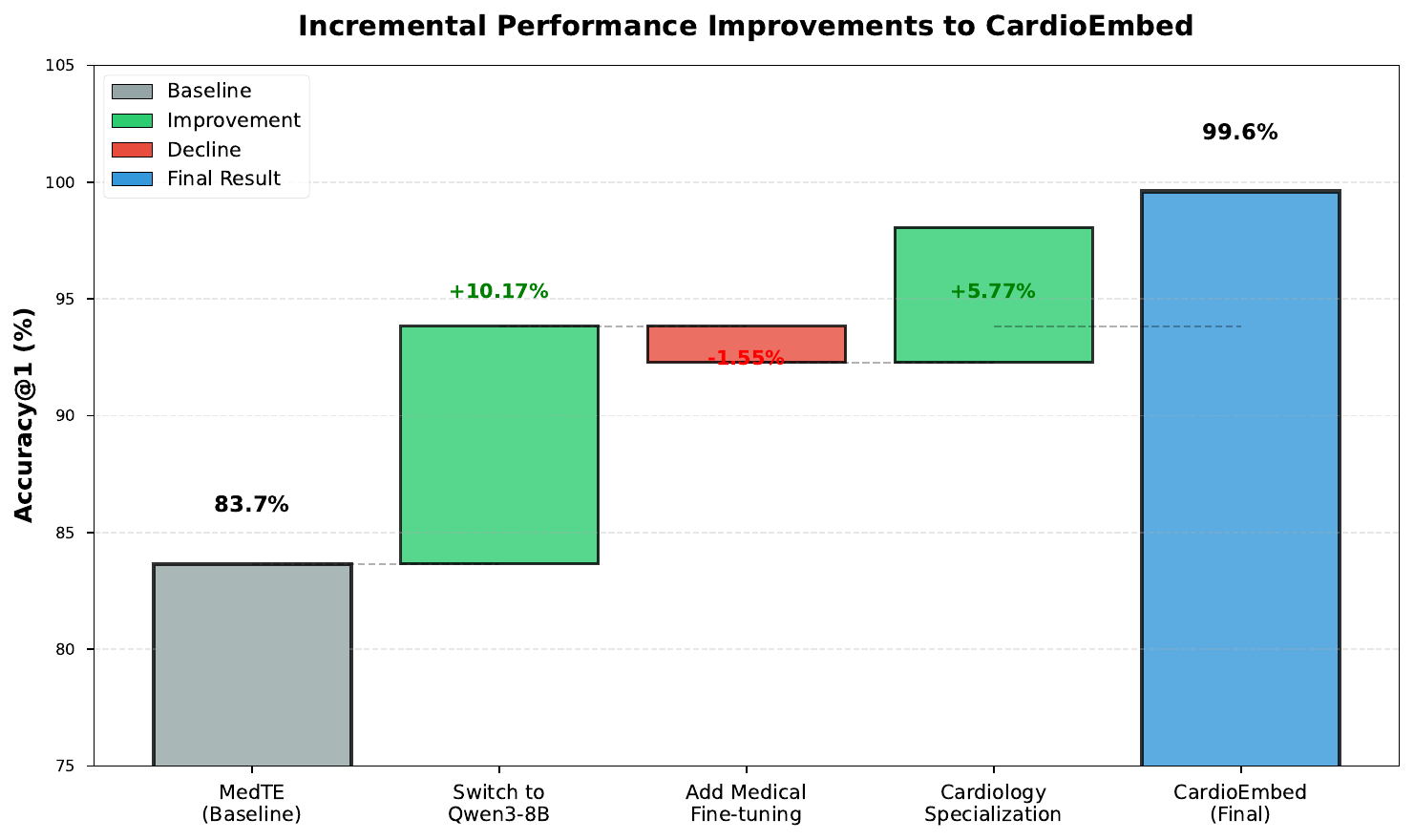}
\caption{Performance improvement waterfall showing incremental contributions to CardioEmbed's final accuracy (higher is better). Starting from MedTE baseline (83.66\%), switching to Qwen3-8B foundation model provided +10.17\% gain, while cardiology-specific fine-tuning added +5.77\%, achieving final performance of 99.60\%. The waterfall visualization demonstrates that both foundation model selection and domain specialization contribute substantially to the final outcome.}
\label{fig:improvement_waterfall}
\end{figure}

Figure \ref{fig:improvement_waterfall} illustrates the incremental contributions to CardioEmbed's final performance. Starting from MedTE's baseline of 83.66\% accuracy, switching to the Qwen3-8B foundation model provided a substantial +10.17 percentage point gain, demonstrating the importance of strong pre-trained language models. Subsequently, domain-specialized cardiology training contributed an additional +5.77 percentage points, bringing final performance to 99.60\%. This decomposition reveals that both foundation model quality and domain-specific fine-tuning contribute substantially to the final outcome.

\begin{figure}[H]
\centering
\includegraphics[width=0.8\textwidth]{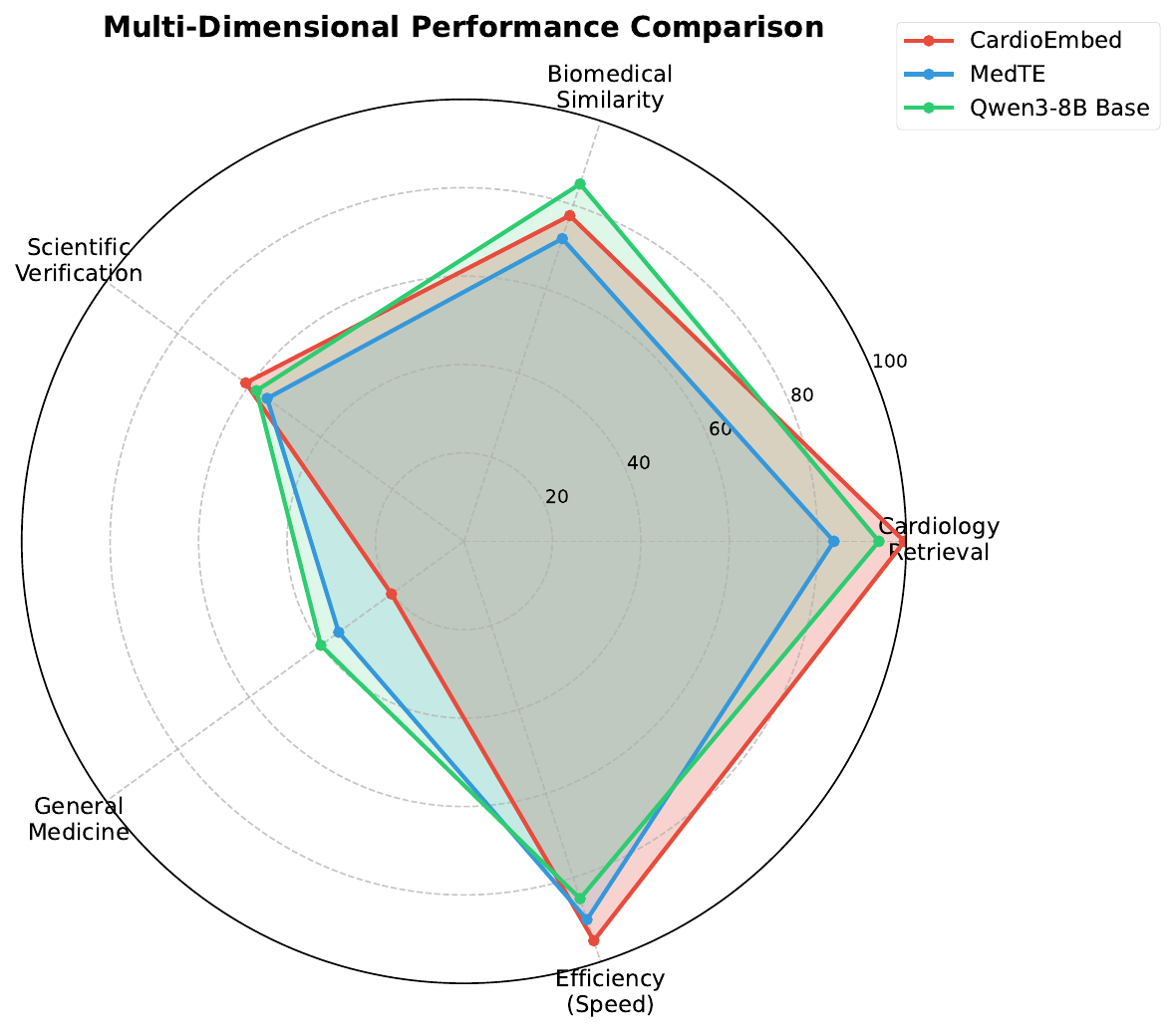}
\caption{Multi-dimensional performance comparison across five key capabilities. CardioEmbed (red) excels in cardiology-specific retrieval (99.6\%) and maintains strong biomedical similarity (77.5\%) and scientific verification (61.0\%) performance. The radar chart illustrates the intentional trade-off: depth in cardiology versus breadth across general medicine. MedTE (blue) shows more balanced but lower overall performance, while base Qwen3-8B (green) demonstrates strong general capabilities with moderate domain-specific performance.}
\label{fig:radar_performance}
\end{figure}

The trade-offs inherent in domain specialization become apparent when examining performance across multiple dimensions. Figure \ref{fig:radar_performance} presents a multi-dimensional comparison of CardioEmbed, MedTE, and the base Qwen3-8B model across five key capabilities: cardiology-specific retrieval, biomedical similarity, scientific verification, general medicine coverage, and computational efficiency. CardioEmbed demonstrates exceptional strength in cardiology-specific tasks while maintaining competitive performance on related biomedical domains, illustrating the depth-versus-breadth paradigm that characterizes specialized embedding models.

MTEB medical benchmark results demonstrate that CardioEmbed maintains strong performance on broader medical domains despite specialized training. The model achieves 0.77 Spearman correlation on BIOSSES biomedical similarity and 0.61 NDCG@10 on SciFact scientific verification, indicating successful transfer to related biomedical domains. However, performance on NFCorpus (0.20 NDCG@10) illustrates the expected specialization trade-off. This performance pattern (excellent on cardiology-specific tasks, strong on biomedical similarity, moderate on general medical retrieval) reflects the intentional design choice to prioritize depth of cardiology knowledge over breadth of general medical coverage.

A notable finding is that the base Qwen3-8B model (without any medical fine-tuning) outperformed all existing medical-specialized models, achieving 93.83\% accuracy. This unexpected result suggests that modern foundation models possess remarkably strong general language understanding capabilities. However, CardioEmbed's domain-specialized training still provides substantial additional value, achieving near-perfect performance that significantly exceeds even this strong foundation.

\subsection{Clinical Implications}

Given the substantial mortality and economic burden of cardiovascular disease (accounting for one-third of global deaths and over \$400 billion in annual U.S. costs), improving clinical decision support through specialized embeddings represents a meaningful contribution to addressing this public health challenge. The substantial performance improvements demonstrated by CardioEmbed have direct implications for clinical cardiology practice. Existing medical information retrieval systems frequently struggle with cardiology-specific queries involving interventional procedures (e.g., transcatheter aortic valve replacement complications), device specifications (e.g., dual-chamber pacemaker indications), and specialized imaging protocols (e.g., stress echocardiography with dobutamine), precisely the knowledge domains where CardioEmbed's textbook-based training provides superior semantic understanding.

Integration of specialized embeddings like CardioEmbed could improve several clinical applications. First, clinical decision support systems querying cardiovascular knowledge bases would retrieve more relevant procedural guidance and treatment protocols. Second, differential diagnosis systems could better match patient presentations to rare cardiovascular conditions by capturing the nuanced clinical reasoning patterns present in textbooks. Third, clinical trial matching platforms could more accurately identify eligible cardiology patients by understanding complex inclusion criteria involving cardiac physiology and procedural history. Fourth, point-of-care literature search for practicing cardiologists would return more relevant results for procedural questions compared to research-focused retrieval systems.

However, clinical deployment requires careful validation. While CardioEmbed demonstrates superior semantic retrieval, it does not perform clinical reasoning, assess evidence quality, or verify factual correctness. Integration into high-stakes clinical workflows would require combination with clinical reasoning frameworks, real-time evidence appraisal systems, and appropriate human oversight. Prospective evaluation in realistic clinical settings remains necessary to demonstrate practical utility and identify potential failure modes before widespread adoption.

\subsection{Limitations}

Several limitations warrant consideration. First, evaluation focused on English-language textbooks and benchmarks; generalization to other languages requires additional validation. Second, the cardiac-specific evaluation reflects a limited sample of potential clinical applications; deployment in actual clinical workflows would provide more definitive assessment. Third, the model has not been evaluated for biases that may be present in the training textbooks. Fourth, while the model captures semantic relationships well, it does not explicitly reason about clinical correctness or safety, requiring integration with clinical reasoning frameworks for high-stakes applications.

\subsection{Future Directions}

Several promising directions warrant investigation. First, the approach should be extended to other medical domains beyond cardiology, including oncology, neurology, radiology, and other specialties, to validate the generalizability of textbook-based domain specialization. Second, expanding the cardiology corpus with additional textbooks covering underrepresented subspecialties (interventional cardiology, electrophysiology, cardiac imaging) would further improve coverage. Third, incorporating multimodal content such as procedural videos and medical imaging would enable richer semantic representations. Fourth, conducting prospective studies evaluating performance in real clinical workflows would provide definitive evidence of practical utility. Finally, investigating optimal training strategies for multi-domain models that maintain specialization while enabling knowledge transfer across related medical fields represents an important theoretical challenge.

\section{Conclusion}

This work demonstrates that domain-specialized embedding models trained on comprehensive clinical textbooks provide superior performance for cardiology-specific semantic tasks compared to models trained exclusively on research literature. The findings highlight a meaningful research-practice gap in medical knowledge representation and demonstrate an effective approach to bridging it through targeted training on clinical textbook content. CardioEmbed achieves substantial improvements on cardiac-specific semantic similarity and information retrieval while maintaining strong general language understanding capabilities. These results suggest that specialized embedding models trained on clinical textbooks represent a promising direction for developing natural language processing systems that better serve the needs of clinical practice.

\section*{Acknowledgments}

The authors thank Dr. Svetlana Barbarash, MD, for contributions to cardiology through clinical practice and research.

The authors acknowledge computational resources provided by NVIDIA H100 GPU infrastructure and the open‑source community for the Qwen3, HuggingFace Transformers \cite{wolf2020huggingface}, and MTEB frameworks.

\section*{Data and Code Availability}

Training data consists of copyrighted cardiology textbooks and cannot be publicly shared. The trained CardioEmbed model weights are publicly available on HuggingFace at \url{https://huggingface.co/richardyoung/CardioEmbed}. Training and evaluation code is available on GitHub at \url{https://github.com/ricyoung/CardioEmbed}. Evaluation datasets (BIOSSES, SciFact, NFCorpus) are publicly available through the MTEB benchmark framework.

\section*{Competing Interests}

The authors declare no competing interests.

\section*{Author Contributions}

R.J.Y. conceived the study, developed the model, conducted experiments, and wrote the manuscript. A.M.M. provided clinical cardiology expertise and reviewed the manuscript.

\bibliographystyle{plain}
\bibliography{bibliography/references}

\end{document}